\documentclass[lettersize,journal,twoside]{IEEEtran}
\usepackage{amsmath,amsfonts}
\usepackage[ruled, lined, longend, linesnumbered]{algorithm2e}
\usepackage{array}
\usepackage[caption=false,font=normalsize,labelfont=sf,textfont=sf]{subfig}
\usepackage{textcomp}
\usepackage{stfloats}
\usepackage{url}
\usepackage{graphicx}
\usepackage{cite}
\usepackage{multirow}

\usepackage[colorlinks=true, citecolor=blue, linkcolor=blue]{hyperref} 
\usepackage{booktabs}       
\usepackage{amssymb}       
\usepackage{flushend}

\begin{document}

\title{FedNIA: Noise-Induced Activation Analysis for Mitigating Data Poisoning in Federated Learning}

\author{Ehsan~Hallaji,~\IEEEmembership{Member,~IEEE,}
		Roozbeh~Razavi-Far,~\IEEEmembership{Senior Member,~IEEE,}
		and Mehrdad~Saif,~\IEEEmembership{Life Fellow,~IEEE}
  
        \thanks{This work was supported by the Natural Sciences and Engineering Research Council of Canada (NSERC) under funding reference numbers CGSD3-569341-2022 and RGPIN-2021-02968.}
        \thanks{Ehsan Hallaji and Roozbeh Razavi-Far are with the Faculty of Computer Science, University of New Brunswick, Fredericton, NB E3B 5A3, Canada. E-mail: e.hallaji@unb.ca, roozbeh.razavi-far@unb.ca.\protect}
		\thanks{Mehrdad Saif is with the Department of Electrical and Computer Engineering, University of Windsor, Windsor, ON N9B 3P4, Canada. E-mail: msaif@uwindsor.ca.\protect}}

\markboth{IEEE Transactions on Knowledge and Data Engineering, September 2026}%
{Hallaji \MakeLowercase{\textit{et al.}}: FedNIA: Noise-Induced Activation Analysis for Mitigating Data Poisoning in Federated Learning}

\maketitle

\begin{abstract}
 Federated learning systems are increasingly threatened by data poisoning attacks, where malicious clients compromise global models by contributing tampered updates. Existing defenses often rely on impractical assumptions, such as access to a central test dataset, or fail to generalize across diverse attack types, particularly those involving multiple malicious clients working collaboratively. To address this, we propose Federated Noise-Induced Activation Analysis (\texttt{FedNIA}), a novel defense framework to identify and exclude adversarial clients without relying on any central test dataset. \texttt{FedNIA} injects random noise inputs to analyze layerwise activation patterns in client models leveraging an autoencoder that detects abnormal behaviors indicative of data poisoning. \texttt{FedNIA} can defend against diverse attack types, including sample poisoning, label flipping, and backdoors, even in scenarios with multiple attacking nodes. Experimental results on non-iid federated datasets demonstrate its effectiveness and robustness, underscoring its potential as a foundational approach for enhancing the security of federated learning systems.
\end{abstract}


\section{Introduction}
\label{sec:intro}
\IEEEPARstart{F}{ederated} Learning (FL) redefines conventional centralized data processing by enabling training across decentralized devices \cite{pmlr-v54-mcmahan17a}. This collaborative methodology aggregates model parameters derived from local training at the network's edge, emphasizing the advantages of local computation. By doing so, FL ensures user privacy by restricting the server's access to local data. The decentralized nature of FL yields significant benefits, including diminished data transfer requirements, enhanced privacy preservation, and the capability to harness diverse datasets from various clients. Despite these advantages, FL is not immune to security threats. Particularly vulnerable to poisoning attacks, FL faces challenges in maintaining the integrity of the collaborative learning process \cite{NEURIPS2020_b8ffa41d, pmlr-v97-bhagoji19a, 247652, 11278062}. In light of the pressing importance of this threat, this paper focuses on fortifying FL against the subtle threat of data poisoning attacks, thereby ensuring the reliability and robustness of the decentralized learning framework.

Data poisoning attacks in FL primarily aim to manipulate the sample distribution and label distribution in both targeted and untargeted manners. These attacks occur on the malicious client side, where adversaries alter the training data before contributing updates to the global model. If undetected or not properly addressed, a malicious update has the potential to contaminate the global model, leading to the spread of the attack into other clients' models.

In untargeted schemes, the goal is to generally degrade the overall performance of the global model \cite{9833647, NEURIPS2022_e2ef0cae}. This type of attack introduces changes to samples or their labels and compromises the model's accuracy across various classes, making it less reliable in its predictions. In targeted data poisoning attacks, the focus is on specific classes or injecting triggers to produce forged predictions upon triggering the system \cite{pmlr-v108-bagdasaryan20a, NEURIPS2023_6ad5d39b, Cao_2022_CVPR}. The aim here is to manipulate the model's behavior strategically, introducing biases or vulnerabilities that can be exploited to compromise the integrity of predictions related to particular classes or conditions. Addressing and thwarting these data poisoning techniques is of crucial importance to maintain the trustworthiness and effectiveness of FL in decentralized environments.

Addressing data poisoning challenges in FL involves exploring various solutions, mainly categorized into anomaly detection, adversarial training, robust aggregation, and the utilization of regularization techniques. Anomaly detection methods have been proven effective against untargeted attacks but may fall short in eliminating targeted ones, often requiring a test dataset on the server for comparison, which may not always be available \cite{li2019abnormal, li2020learning, 247652, 291249, cao2022fltrust}. Adversarial training assumes knowledge of the attack distribution and may struggle when confronted with new attack patterns. Regularization techniques, such as pruning and dropout, provide partial robustness but do not offer comprehensive protection against attacks. Robust aggregation, while enhancing model resilience, often results in lower performance compared to federated averaging (\verb+FedAvg+) in the absence of attacks. Moreover, it is important to note that the majority of the available security solutions are typically effective against specific groups of attack. For instance, several research efforts are dedicated to addressing backdoors in FL \cite{pmlr-v139-xie21a, NEURIPS2023_2376f25e, 280048, 10577215}. A framework that offers robust protection against all types of data poisoning attacks and can handle several attacking nodes remains rare. Available security solutions, however, are not limited to the mentioned groups, as the landscape of FL security is continually evolving, and novel approaches may emerge to counter emerging threats. The key lies in understanding the strengths and limitations of each approach and striving for a balanced and comprehensive defense strategy.

In pursuit of a comprehensive approach to mitigate data poisoning in FL, a novel client update verification mechanism, called \verb+FedNIA+, is proposed that can filter out potentially malicious updates from the aggregation process. This mechanism, combined with any aggregators such as \verb+FedAvg+ enables robust aggregation. This approach eliminates the need for having a private test dataset on the server and analyzes changes in the activation outputs of network layers. Unlike other robust aggregators, the proposed method maintains a performance level similar to the baseline when no attack is present, striking a balance between resilience and performance in the absence of threats. Moreover, this mechanism is robust against various mechanisms of sample poisoning, label flipping, and backdoor attacks, covering the primary categories of data poisoning attacks. Notably, our approach is able to mitigate collaborative poisoning attacks even with high ratios of malicious to benign clients. 
Compared to existing anomaly detection and trust-based defenses, which primarily operate in the parameter space by directly analyzing client updates or their similarity to reference signals \cite{291249,cao2022fltrust,10.1145/2991079.2991125}, the proposed approach differs in both representation and mechanism. \verb+FedNIA+ evaluates client updates through their induced activation behavior when applied to the global model, effectively shifting the detection process from parameter space to a functional representation. In addition, it captures activation dynamics under controlled perturbations, rather than relying on static similarity or validation-based evaluation.

The remainder of this paper is organized as follows. Section \ref{sec:related_work} reviews the related works. Section \ref{sec:threat_model} formulates the threat models used in this work. Section \ref{sec:methodology} presents the proposed methodology. Experimental results are reported and analyzed in Section \ref{sec:results}. Finally, the paper is concluded in Section \ref{sec:conclusion}.

\section{Related Works}
\label{sec:related_work}
The literature on FL robustness against poisoning attacks can be categorized into two main streams: server-side and client-side defense \cite{pmlr-v202-zhu23j, NEURIPS2021_692baebe, pmlr-v139-xie21a, NIPS2017_f4b9ec30, 291249}. These two can be used simultaneously, as they secure different ends of a FL network. Server-side defense, which is the topic of interest in this work, is often carried out using anomaly detection or tailored aggregation rules that make the central aggregator robust against poisoning attacks. Here, we refer to such methods as robust aggregation. Moreover, a number of studies show that partial robustness against poisoning attacks can be obtained using regularization techniques and Differential Privacy (DP) \cite{sun2019really, naseri2022local}. In the following overview, we will concisely go through both domains and underscore their relevance to our research.

\subsection{Poisoning Attacks}
Poisoning attacks generally fall under model poisoning and data poisoning categories \cite{10420449, 9945997}. In data poisoning, the attacker trains a local model using corrupted data samples to satisfy a malicious objective. A common objective in data poisoning is increasing the overall classification error (i.e., untargeted setting) or deteriorating the classification performance on specific classes (i.e., targeted setting). Data poisoning can be accomplished by corrupting the samples or their labels. When corrupting the label, these attacks are referred to as label flipping (i.e., also known as label poisoning) \cite{HALLAJI2023110384}. Data poisoning can be also in the form of backdoor attacks, where specific trigger patterns are injected into the local data \cite{9802938}. This targeted attack ideally does not affect the classification performance and only misclassifies samples that contain the specified trigger.

In model poisoning, the attacker controls the global model by manipulating the objective function \cite{10816078}. If the local model is trained using a malicious objective on valid data, the obtained parameters can poison the global model \cite{10420449}. As an example, the malicious objective can include an additional term to the global objective function that penalizes sensitivity to the corrupted data \cite{pmlr-v97-bhagoji19a}. The same strategy can be employed to decrease the global classification performance in an untargeted manner \cite{9210138}. Another study in \cite{9293091} uses pointwise Gaussian noise to poison gradients. Although model poisoning can be performed by directly manipulating the gradients, data poisoning is also used as a tool to perform model poisoning \cite{DBLP:journals/corr/abs-2202-02817}. This further signifies the importance of mitigating data poisoning attacks in FL.

The aforementioned poisoning attack can be orchestrated using multiple malicious clients. This setting is often referred to as a Byzantine attack, where multiple nodes collaboratively launch an attack in a FL network \cite{10878290}. When a single client creates several fake nodes to satisfy the same objective, a Sybil attack takes place \cite{259745}.

\subsection{Robust Aggregation}
Several robust aggregation techniques have been introduced to tackle data poisoning in FL, while aiming at maintaining the FL performance. A group of these algorithms detects suspicious updates and reduces their contribution weight to the aggregation process on the server. For instance, the aggregator in \cite{pmlr-v80-yin18a} calculates the median or coordinate trimmed mean of local updates prior to generating the global update. Another group of algorithms finds clusters of clients and sets malicious clients apart from benign users so that the suspicious users do not participate in the aggregation process. As an example, FoolsGold \cite{259745} combats Sybil attacks by adjusting the learning rates of local models based on a contribution similarity. This method effectively identifies Sybil groups, when they are present. Nonetheless, it is prone to mistakenly flag benign participants and deteriorate the training performance. While this method relies on the similarity of malicious users, other approaches such as \cite{NIPS2017_f4b9ec30, pmlr-v80-mhamdi18a} take the correlation of benign users into account. In addition, statistical methods such Robust Federated Aggregation (\verb+RFA+), which use the median of updates, have been shown effective in enhancing attack robustness \cite{9721118}. Robust aggregation with adaptive clipping (\verb+AdaClip+) is performed by zeroing out extremely large values for robustness to data corruption on clients, and adaptively clipping in the L2 norm to the moderately high norm for robustness to outliers \cite{NEURIPS2021_91cff01a}.  \verb+FLAME+ clusters client updates to identify suspicious groups, removing outliers, and applying adaptive clipping and noise injection to limit the impact of malicious updates \cite{280048}.

\subsection{Differential Privacy}
While DP has been primarily considered as a defense against inference attacks, several studies show that it can also be effective in making the FL model more robust against poisoning attacks \cite{10.5555/3489212.3489304, NEURIPS2020_fc4ddc15, ijcai2019p657, naseri2022local, sun2019really, 9669031}. An adaptive version of DP for FL is presented in \cite{NEURIPS2021_91cff01a} (\verb+AdaDP+) that clips the norm at a specified quantile of the update norm distribution similar to \verb+AdaClip+. The value at the quantile is calculated in real-time with DP.

\subsection{Anomaly Detection}
\label{sec:anomaly}
Anomaly detection techniques are actively employed to identify and prevent malicious updates from compromising systems. In FL, these methods can be employed on the server side to detect potential threats. FL-based anomaly detectors often utilize a central test data to evaluate updates and filter out those that are detrimental or neutral to the global model.

For instance, a detection method described in \cite{10.1145/2991079.2991125} groups users into benign and dangerous categories by clustering clients based on the information received to identify malicious updates. Another technique uses a distance metric for several participants to track updates' drift \cite{NIPS2017_f4b9ec30}. To find anomalous updates from participants, a different method creates low-dimensional model weight surrogates \cite{DBLP:journals/corr/abs-1910-09933}. Among other updates, those that support the objective function are chosen using an outlier detection strategy. Examples of anomaly detection in FL include several methods outlined in recent studies \cite{247652, li2020learning}.

Prior work has explored activation-based analysis for anomaly detection in deep learning. For instance, \cite{10.1007/978-3-030-67661-2_5} and its extensions analyze hidden-layer activation patterns to distinguish normal and anomalous inputs, while DA3D leverages activation representations within adversarial autoencoder frameworks to detect previously unseen anomalies \cite{9892896}. Similarly, subset scanning approaches examine anomalous patterns in autoencoder activations and reconstruction signals \cite{ijcai2020p122}, and ALAD formulates anomaly detection through activation learning in time-series settings \cite{10663916}. Despite their differences in modeling and objectives, these methods share a common paradigm of operating in centralized settings and analyze activations induced by input data within a single model. 

In contrast, our work addresses a fundamentally different problem in FL, where the goal is to identify anomalous client updates prior to aggregation. Rather than analyzing activations induced by data samples, the proposed approach evaluates model responses driven by client-provided updates, without requiring access to local data. Moreover, it characterizes activation dynamics under controlled perturbations, enabling detection based on the stability of activation behavior rather than static representations. From the perspective of FL defense taxonomies, this places the proposed method as a server-side anomaly detection approach that filters malicious updates prior to aggregation, without relying on validation data, while remaining robust to non-i.i.d. client distributions \cite{10423783}.

\section{Threat Models}
\label{sec:threat_model}
Poisoning attacks can be conducted in various settings with different objectives. In this work, we consider three main categories of these attacks, namely sample poisoning, label flipping, and backdoors \cite{247652, Ozdayi_Kantarcioglu_Gel_2021, 9721118}. For the sake of simplicity, the formulation is presented assuming a single attacker; however, in our experimental setting, multiple adversaries can simultaneously launch each attack.

\subsection{Sample Poisoning}
Let $X = \{x_1, x_2, \dots , x_m\}$ and $Y = \{y_1, y_2, \dots, y_m\}$ represent benign training samples and labels at time $t$, where $X \in \mathbb{R}^{n}$ and $Y \in \mathbb{N}$. $m$ and $n$ denote the number of samples and dimensions, respectively. A dataset $\mathcal{D} = \{(x_i, y_i)\}$ is defined as a set of tuples drawn from $X$ and $Y$, where $1\leq i\leq m$. The attacker aims to obtain a poisoned dataset $\tilde{\mathcal{D}}$ by injecting a set of malicious samples $\Delta x$ into $\tilde{\mathcal{D}}$ to minimize the model's performance. Under an untargeted setting, sample poisoning is carried out as:
\begin{equation}
    \tilde{\mathcal{D}} = \{(x_i + \Delta x, y_i) \mid P(\Delta x \neq 0) = \gamma \}_{i=1}^m,
\end{equation}
where $\gamma$ is a parameter determining the ratio of poisoned samples in $\tilde{\mathcal{D}}$. Under a targeted setting, the attacker distorts the sample data for a specific class $c$:
\begin{equation}
    \tilde{\mathcal{D}} = \{(x_i + \Delta x, y_i) \mid P(\Delta x \neq 0 \mid y_i = c) = \gamma \}_{i=1}^m
\end{equation}

\subsection{Label Flipping}
Let $y_i$ represent the true label of a sample $x_i$, and $\hat{y}_i$ denotes the label assigned by the model. The attacker aims to modify the labels of a subset of the training data to minimize the model's performance. In the untargeted scheme, the attacker randomly flips labels:
\begin{equation}
    \tilde{\mathcal{D}} = \{( x_i, \hat{y}_i) \mid P(\hat{y}_i \neq y_i) = \gamma\}_{i=1}^m
\end{equation}
In the targeted scheme, the attacker specifically targets a certain class $c$, switching its labels to other classes:
\begin{equation}
    \tilde{\mathcal{D}} = \{(x_i, \hat{y}_i) \mid P(\hat{y}_i \neq y_i) = \gamma \land y_i = c\}_{i=1}^m
\end{equation}

\subsection{Backdoor}
The attacker aims to modify the training data distribution to include backdoor triggers $\epsilon$ to the samples of class $c$ so that when the trigger is activated or observed in samples, it will be predicted as $y_{backdoor}$ to manipulate the model's behavior:
\begin{equation}
    \tilde{\mathcal{D}} = \{(x_i + \epsilon, y_{backdoor}) \mid y_i = c\}_{i=1}^m
\end{equation}
$\epsilon \in \mathbb{R}^n$, and $y_{backdoor} \neq c$ in the above formulation. Backdoors are targeted attacks by nature. We consider two backdoor mechanisms: pixel backdoor and injecting specific noise patterns into the data.

In our simulations, attackers have white-box access to local models, meaning they can see the architecture of the local model but have no knowledge of the aggregation structure on the server. This implies that the attackers can inspect the internal workings of individual models but do not have information about how these models are combined or aggregated at a higher level.

\section{Noise Induced Activation Analysis}
\label{sec:methodology}
In this section, the design of the proposed method, \verb+FedNIA+, and the motivation behind it are explained. Moreover, we explain how other FL system components interact with the proposed approach. We begin by outlining the intuition behind the proposed approach, followed by its detailed formulation.

\subsection{Motivation}
As reviewed in Section \ref{sec:anomaly}, prior work has shown that hidden-layer activations can encode useful signals for anomaly detection, where deviations in activation patterns reflect irregularities in model behavior. Building on this intuition, we posit that client updates may induce distinguishable activation responses under input perturbations. In particular, benign updates tend to produce more consistent activation patterns, while malicious updates may result in less stable responses, especially under the assumption that the majority of client updates are benign. By introducing controlled noise and analyzing these activation dynamics, the proposed method aims to identify anomalous client updates prior to aggregation.

To provide a functional interpretation of this assumption, let
$\phi(W)=\mathbb{E}_{Z}[A(W,Z)]$ denote the expected activation
representation of a model with parameters $W$ under the random probe
distribution. For a client model
$W_i^t = W_G^t + \Delta W_i^t$, a first-order expansion around the
global state gives
\begin{equation}
\phi(W_i^t)-\phi(W_G^t)
\approx
J_{\phi}(W_G^t)\Delta W_i^t,
\end{equation}
where $J_{\phi}$ is the Jacobian of the activation representation with
respect to the model parameters. Consequently,
\begin{equation}
\left\|
\phi(W_i^t)-\phi(W_G^t)
\right\|_2^2
\approx
(\Delta W_i^t)^{\top}
J_{\phi}^{\top}J_{\phi}
\Delta W_i^t.
\end{equation}
Since $J_{\phi}^{\top}J_{\phi}$ is positive semidefinite, the
activation representation induces a local functional measure of a
client update, weighting parameter changes according to their effect
on the model's internal behavior rather than their magnitude in
parameter space alone. \verb+FedNIA+ builds on this principle by using
reconstruction error to identify deviations from the activation
behavior represented by the global model. This provides a theoretical basis for analyzing client updates in activation space.

\subsection{Client Models} 
Based on the aforementioned intuition, we next formalize the client-side model and update generation process. Consider a set of $k$ benign clients, denoted by $C=\{c_1, c_2, \dots, c_k\}$. A client model $M$ is formally defined as:
\begin{equation}
M(W_i, x):  x, h_1^M, h_2^M, \dots, h_L^M, \hat{y}
\end{equation}
for client $c_i$ with network weights $W_i$ and $L$ hidden layers $h_l$. Indicating the FL time-steps with $t$, we denote trained model weights and the local dataset at each iteration with $W_i^t$ and $\mathcal{D}_i^t$, respectively. At each time step, $M$ is first initialized with weights of the global model $W_G^t$, which we refer to as the global state. Then, $M$ is trained on the local data $\mathcal{D}_i^t$ to obtain $W_i^t$ as $M(W_G^t, \mathcal{D}_i^t)\rightarrow W_i^t$. Afterward, the locally trained model parameters $W_i^t$ are communicated to the server. For conciseness, these submitted model parameters are referred to as client updates throughout the remainder of the paper. This process is also shown in Algorithm \ref{alg:client}.

\subsection{Malicious Clients} 
We now extend the clients' formulation to account for adversarial behavior in the presence of malicious clients. Algorithm \ref{alg:client} also details the condition where malicious clients $\tilde{C}=\{\tilde{c}_1, \tilde{c}_2, \dots, \tilde{c}_r\}$ can initiate a poisoning attack on the FL process by training the model on poisoned data $\tilde{\mathcal{D}}_i^t$ and obtain $M(W_G^t, \tilde{\mathcal{D}}_i^t) \rightarrow \tilde{W}_i^t$. If not eliminated by the aggregator, sending $\tilde{W}_i^t$ to the server will poison the next global state $W_G^{t+1}$ and corrupt all $W_i^{t+1}$ as a result. Here, the number of malicious clients is assumed to be $1 \leq r < k$. This is because when $r\geq k$, malicious clients constitute at least half of the participating clients, so the benign-majority assumption no longer holds. In particular, $r=k$ corresponds to exactly 50\% malicious clients, while $r>k$ corresponds to a malicious majority.

\begin{algorithm}[t]
\caption{Clients}
\label{alg:client}
\SetKwInput{Option}{Option}
\Option{Benign or malicious}
\KwIn{Local datasets $\mathcal{D}$ (or malicious dataset $\tilde{\mathcal{D}}$), global state $W_G$}
\KwOut{Local updates $W$ (or malicious updates $\tilde{W}$)}
 \For{$t \in [0, T]$}{
   \Switch{Option}{
   \Case{Malicious}{
   \For{client $i \in [1, r]$ in parallel}{
      Train $M(W_G^t, \mathcal{\tilde{D}}_i^t)$\\
      Create update $\tilde{W}_i^t \gets M(W_G^t, \mathcal{\tilde{D}}_i^t)$\\
   }}
   \Case{Benign}{
   \For{client $i \in [1, k]$ in parallel}{
      Train $M(W_G^t, \mathcal{D}_i^t)$\\
      Create update $W_i^t \gets M(W_G^t, \mathcal{D}_i^t)$\\
   }}
   }
 }
\end{algorithm}

\begin{algorithm}[t]
\caption{\texttt{FedNIA}}
\label{alg:server}
\KwIn{Client weights $W_i^t$}
\KwOut{Next global State $W_G^{t+1}$}
\SetKwInput{Initialization}{Initialization}
\Initialization{Initialize the global model $W_G^0$}
 \For{$t \in [0, T]$}{
  Generate random inputs $Z^t$\\
  \For{$W_i^t$ in parallel}{
   Get layer activations $A_i^t \gets M(W_i^t, Z^t)$\\
   Estimate the average activations for each $c_i$: $\bar{A}_i^t = \frac{1}{\nu}\sum_{j=1}^\nu A_{i(j)}^t$
  }
  Get layer activations $A_G^t \gets M(W_G^t, Z^t)$\\
  Compute averaged layer activations $\bar{A}_G^t = \frac{1}{\nu}\sum_{j=1}^{\nu}A_{G(j)}^t$\\
  Train autoencoder $AE$ on $\bar{A}_G^t$ w.r.t. (\ref{eq:loss})\\
  \For{$W^t_i$}{
   Feed $\bar{A}_i^t$ to the detector:  $\hat{A}_i^t \gets AE(\bar{A}_i^t)$\\
   Compute reconstruction error as: $e_i^t = \sqrt{\frac{1}{k+r}\parallel \bar{A}_i^t - \hat{A}_i^t \parallel^2}$\\
  }
  Calculate threshold $\tau^t$ using (\ref{eq:tau})\\
  Filter updates $\Omega^t = \{W^t_1, W^t_2, \dots, W_k^t\}$, where $W^t_i = \varnothing$ if $e_i^t > \tau^t$\\
  $W_G^{t+1} \gets$ Aggregate updates FedAvg$(\Omega^t)$\\
 }
\end{algorithm}

\subsection{Server} Given the presence of potentially malicious updates, the server must evaluate incoming client models before aggregation. In this process, the FL server will receive a number of updates $W_i^t$ from clients. For simplicity, all $k+r$ clients are assumed to participate in each FL round. Accordingly, the server receives one submitted client model $W_i^t$ for each $i\in\{1,\dots,k+r\}$. At this step, the server aims to evaluate $W_i^t$ to filter out potentially malicious updates. Since different mechanisms of poisoning attack are likely to affect the model $M$ differently, designing a detection model that handles a spectrum of attacks requires careful analysis of $W_i^t$ at each layer for all incoming updates. Nevertheless, $W_i^t$ is often high-dimensional with millions of values included in weight matrices. Therefore, resorting to anomaly detection principles to directly process $W_i^t$ may not be efficient. A better alternative would be reconstructing $M(W_i^t, \mathcal{D}_{test})$ on the server so that the model could be evaluated on a test dataset $\mathcal{D}_{test}$ based on the resulting outputs. Nonetheless, access to client data is prohibited, and considering central $\mathcal{D}_{test}$ requires the assumption that the data distribution remains unchanged in time, which is unrealistic in most cases. Server-side operations are explained in Algorithm \ref{alg:server}.

\begin{figure*}[t]
    \centering
    \includegraphics[width=\textwidth, trim={0.6cm 0.5cm 0.6cm 0.6cm}, clip]{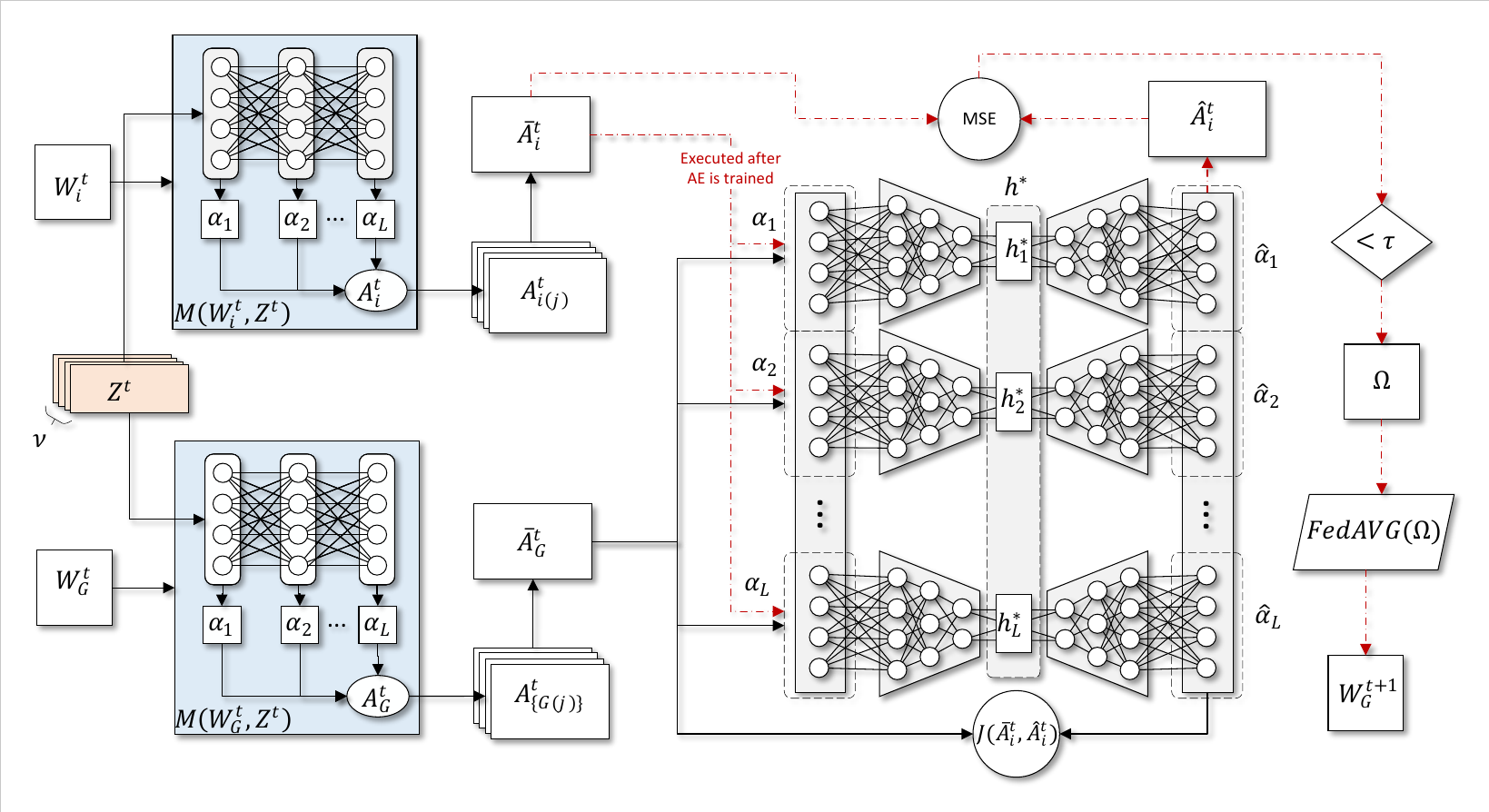}
    \caption{Block diagram of \texttt{FedNIA} on the server. Dashed red lines indicate steps of the algorithm that should be undertaken once all steps specified with black solid lines are completed.}
    \label{fig:diagram}
\end{figure*}

\subsection{Detection Model}

We now describe the overall detection pipeline as summarized in Algorithm \ref{alg:server}. This process consists of four main steps: (i) activation extraction using noise inputs, (ii) autoencoder-based reconstruction, (iii) anomaly scoring via reconstruction error, and (iv) threshold-based filtering. In this process, each layer $l$ of the model $M$ produces a vector of activation values, denoted as $\alpha_l$, where the length of $\alpha_l$ equals the number of neurons in that layer. For a network with $L$ layers, $A_i^t$ represents the concatenation of these vectors $\alpha_1, \alpha_2, \dots, \alpha_L$ for client $i$ at time $t$. To enable the comparison between the average of noise-induced (i.e., obtained by feeding noise to the model) client activations $\bar{A}^t_i$, $1\leq i \leq k + r$, and the activations of the global state, $\bar{A}^t_G$, noise input $Z^t$ is also passed to the model $M(W_G^t, Z^t)$. Using noise inputs instead of real data allows the server to analyze model behavior without relying on sensitive client information, while still surfacing model-level inconsistencies. At this stage, $\bar{A}_G^t$ can be used as a reference for comparison with $\bar{A}_i^t$, where $1\leq i \leq k + r$. 
Although the choice of noise distribution may affect activation magnitudes, the proposed method relies on relative differences in activation behavior and is therefore not strongly dependent on the specific distribution. The use of randomness also introduces uncertainty that can hinder adversarial optimization in FL \cite{10538666}.

To enable anomaly detection, we first construct an autoencoder model with sub-networks to learn layer-level activation patterns, as this enables localized reconstruction of behaviors associated with each layer. This design is intended to allow each sub-network to focus on encoding and decoding activation values from its corresponding layer. Next, the encoder operates as follows. Each sub-encoder $E(\alpha_l)$ takes $\alpha_l$ portion of $\bar{A}_i^t$ that corresponds to layer $l$ of $M(W_i^t, Z^t)$ as in the following:
\begin{equation}
    E(\alpha_l): \alpha_l, h^E_{l(1)}, h^E_{l(2)}, \dots, h^*_l
\end{equation}
where $h_{l(j)}^E$ denotes the $j$-th hidden layer of the sub-network $E(\alpha_l)$, and $h_l^*$ is the code layer of $E(\alpha_l)$. The encoder network is then formally defined as:
\begin{equation}
    \operatorname{Encoder}(\bar{A}): \bar{A}, \bigcup_{l=1}^L E(\alpha_l) \mid \alpha_l \in \bar{A}
\end{equation}
Finally, the decoder reconstructs the activation patterns as follows. Correspondingly, sub-decoders $D_l$ and the decoder are formulated as:
\begin{equation}
    D(h^*_l): h^*_l, h^D_{l(1)}, h^D_{l(2)}, \dots, \hat{\alpha}_l
\end{equation}
\begin{equation}
    \operatorname{Decoder}(h^*): \bigcup_{l=1}^L D(h^*_l), \hat{A} \mid h^*_l  \in h^*
\end{equation}
where $\hat{\alpha}$ is the reconstructed activation values of $h_l^M$, and $\hat{A}$ is the estimated vector of all activation values. Also, $h^* = \{h_1^*, h_2^*, \dots h_L^*\}$ is the code layer connected to the sub-networks in the encoder and decoder. The autoencoder is formulated as $AE(\bar{A}) = \operatorname{Decoder}\left(\operatorname{Encoder}(\bar{A})\right)$. The architecture of the trained $AE$ and other components of the server are depicted in Fig. \ref{fig:diagram}. As illustrated in this figure, the autoencoder learns the activation patterns of the global model and evaluates client updates based on reconstruction error.

\subsection{Privacy Considerations}
A potential concern in the proposed framework is that analyzing activation patterns on the server may introduce privacy leakage risks by revealing information about client-specific model behaviors. However, our approach is designed to mitigate this risk in several ways. First, the activations analyzed are generated using random noise inputs ($Z^t$), not client data. These noise-induced activations are decoupled from any meaningful or task-specific information and are not representative of the clients' training data distributions. Moreover, the activations are aggregated over multiple random inputs and across all neurons, further reducing any residual client-specific signal. Consequently, the information they carry is unlikely to reveal sensitive client details.

Second, we note that our design aligns with the standard assumptions in FL, where client-side model weights are communicated to a central server. This practice itself can potentially leak private information, but it operates under the commonly accepted assumption that the server is trusted, which is fundamental to most FL protocols. Our method does not introduce any additional form of client data exposure beyond this existing assumption.

\subsection{Training Loss} Reconstruction error is used as a proxy for anomaly scoring, following common practices in unsupervised detection, where higher errors indicate deviation from learned benign patterns. Minimizing the reconstruction error on the entire $\hat{A}^t$ vector may unevenly weigh layers with more neurons, which can reduce the fidelity of anomaly detection. Given that $|h_l^M|$ is different for $1\leq l \leq L$, the reconstruction error associated with each $\hat{\alpha}_l$ does not equally contribute to the training loss. For example, in a binary classification problem $|\alpha_L|=2$; however, other hidden layers may have hundreds of values. Therefore, the training loss $J$ of the $AE$ network is defined as the average of root mean squared errors of separate layers as follows:
\begin{equation}
    J(\bar{A}^t_i, \hat{A}^t
    _i) = \frac{1}{L}\sum_{l=1}^L \sqrt{\frac{\parallel\alpha_l - \hat{\alpha}_l\parallel^2}{|\alpha_l|}}
    \label{eq:loss}
\end{equation}

\subsection{Filtering Updates}
Using the reconstruction error obtained from the trained autoencoder, client updates are filtered as follows. A client update is considered anomalous if its reconstruction error exceeds a threshold derived from the distribution of errors across all clients. To do so, at each FL round, $AE$ is first trained on $\bar{A}_G^t$. Then, $AE$ encodes and reconstructs $\bar{A}_i^t$ and receives $\bigcup_{i=1}^{k+r} \hat{A}_i^t = AE(\bar{A}_i^t)$. Then, the reconstruction error for each $\bar{A}_i^t$, $1 \leq i \leq k + r$, will be computed as $e_i^t = \sqrt{\frac{1}{k+r}\parallel \bar{A}_i^t - \hat{A}_i^t \parallel^2}$. Once the errors are estimated, $c_i$ corresponding to $e_i^t$ will be filtered based on the threshold $\tau^t$:
\begin{equation}
    \tau^t = \left(\frac{1}{k + r}\sum_{i=1}^{k+r} e_i^t\right) + \lambda \sigma^t
    \label{eq:tau}
\end{equation}
where $\lambda$ is the scaling factor, and $\sigma^t$ denotes the standard deviation of obtained errors.

The parameter $\lambda$ controls the sensitivity of the filtering
rule. Since anomalous models are identified from the upper tail of the
reconstruction-error distribution, the threshold can be interpreted
using Cantelli's inequality, i.e., the one-sided form of Chebyshev's
inequality. For a random variable $E$ with finite mean $\mu$ and
standard deviation $\sigma$,
\begin{equation}
\Pr\left(E-\mu \geq \lambda\sigma\right)
\leq
\frac{1}{1+\lambda^2},
\qquad \lambda > 0.
\end{equation}
Thus, $\mu+\lambda\sigma$ provides a distribution-free upper-tail
criterion, with larger values of $\lambda$ corresponding to less
aggressive filtering. In \verb+FedNIA+, $\mu$ and $\sigma$ are estimated
from the client reconstruction errors; therefore, this bound provides
statistical grounding for the form of the threshold rather than a
guarantee that every model exceeding it is malicious.

Using the defined threshold, the server retains the submitted client models whose reconstruction errors do not exceed the threshold into $\Omega^t = \{W^t_1, W^t_2, \dots, W_k^t\}$ to prevent $\tilde{c}_i \in \tilde{C}$ contribute in the aggregation step:
\begin{equation}
\bigcup_{i=1}^{k+r}  
    \begin{cases}
        \varnothing & \text{if $e_i^t > \tau^t$}  \\
        W_i^t  & \text{otherwise}
    \end{cases}
\end{equation}
The aggregation of $W_i^t$ is performed using the \verb+FedAvg+ algorithm. At the end of round $t$, \verb+FedAvg+$(\Omega^t) = W_G^{t+1}$ estimates the global state for $t+1$, when the explained process will be repeated.

The autoencoder is updated incrementally across rounds using the current global activation patterns. Since anomalous updates are filtered prior to aggregation, they do not contribute to the global model or subsequent autoencoder training. As a result, the learned representation is primarily influenced by benign updates. In cases where some malicious updates bypass the filtering stage, their impact remains limited, as the global model, and consequently the activation patterns, are dominated by the majority of benign client updates.

At round $t$, the global model $W_G^t$ is fixed before the incoming client models are evaluated. Therefore, the client models submitted in the current round cannot influence the reference used for their own assessment. Under the assumed benign-majority setting, $W_G^t$ provides a practical data-free reference for identifying client models that induce abnormal activation behavior.

\subsection{Complexity Analysis}
We now analyze the computational complexity of the proposed method. For the sake of simplicity, we assume all clients are benign and $r=0$. In each FL iteration $t$, \verb+FedNIA+ initially generates $\nu$ random inputs, which has a complexity of $O(\nu)$. Next, obtaining layer activations for each client model involves a forward pass, leading to a complexity of $O(k \nu |W|)$, where $k$ is the number of updates received from clients, and $|W|$ is the total number of weights in the model. Averaging these activations across all clients adds a complexity of $O(k \nu  \eta)$, where $\eta$ is the total number of activations across all layers. Training the autoencoder on the averaged activations $\bar{A}_G^t$ involves $\beta$ epochs with each epoch having a complexity of $O(\eta |\theta|)$, where \(|\theta|\) is the number of parameters in the autoencoder, leading to a total complexity of $O(\beta \eta |\theta|)$ for the training phase. The inference phase, which includes encoding and reconstructing activations, adds $O(k \eta |\theta|)$. Calculating the reconstruction error for each client is $O(k \eta)$, and computing the threshold and filtering updates both contribute a complexity of $O(k)$. Finally, aggregating the filtered updates, akin to \verb+FedAvg+, has a complexity of $ O(k |W|)$. Assuming that $\nu, \eta \ll |W|$ and $|\theta| < |W|$, the dominating terms in the combination of the aforementioned terms, simplify to $O(k |W| + \beta \eta |\theta|)$, indicating that time complexity of $O(\beta \eta |\theta|)$ is added to that of the \verb+FedAvg+.

The additional operations introduced by \verb+FedNIA+ are performed entirely on the server. Therefore, the method does not require additional client-side computation or communication beyond transmitting the locally trained models required by the underlying FL protocol. For fixed noise and detector configurations, the server-side cost scales linearly with the number of participating client models, consistent with the runtime results in Section \ref{sec:runtime}.

\subsection{Model Generalization Across Architectures}
Finally, we discuss the applicability of the proposed approach across different model architectures. The proposed framework is model-agnostic and can be applied to different neural network architectures, as it operates on layer-wise activation patterns rather than specific model structures. While deeper models may increase the dimensionality of activation representations and the associated computational cost, the underlying detection mechanism remains unchanged. In principle, the approach can be extended to more complex architectures, such as Transformer-based models, where intermediate representations can be incorporated into the activation analysis.

\section{Experimental Results}
\label{sec:results}

This section evaluates the proposed method under several scenarios. The evaluation process involves comparing \verb+FedNIA+ with other defense mechanisms usable in the selected attack scenarios. The experimental setup and attack scenarios are reported and analyzed accordingly. 

\begin{figure*}[t]
    \centering
    
    \includegraphics[trim={0.5cm 0.5cm 0.5cm 0.5cm},clip,width=\textwidth]{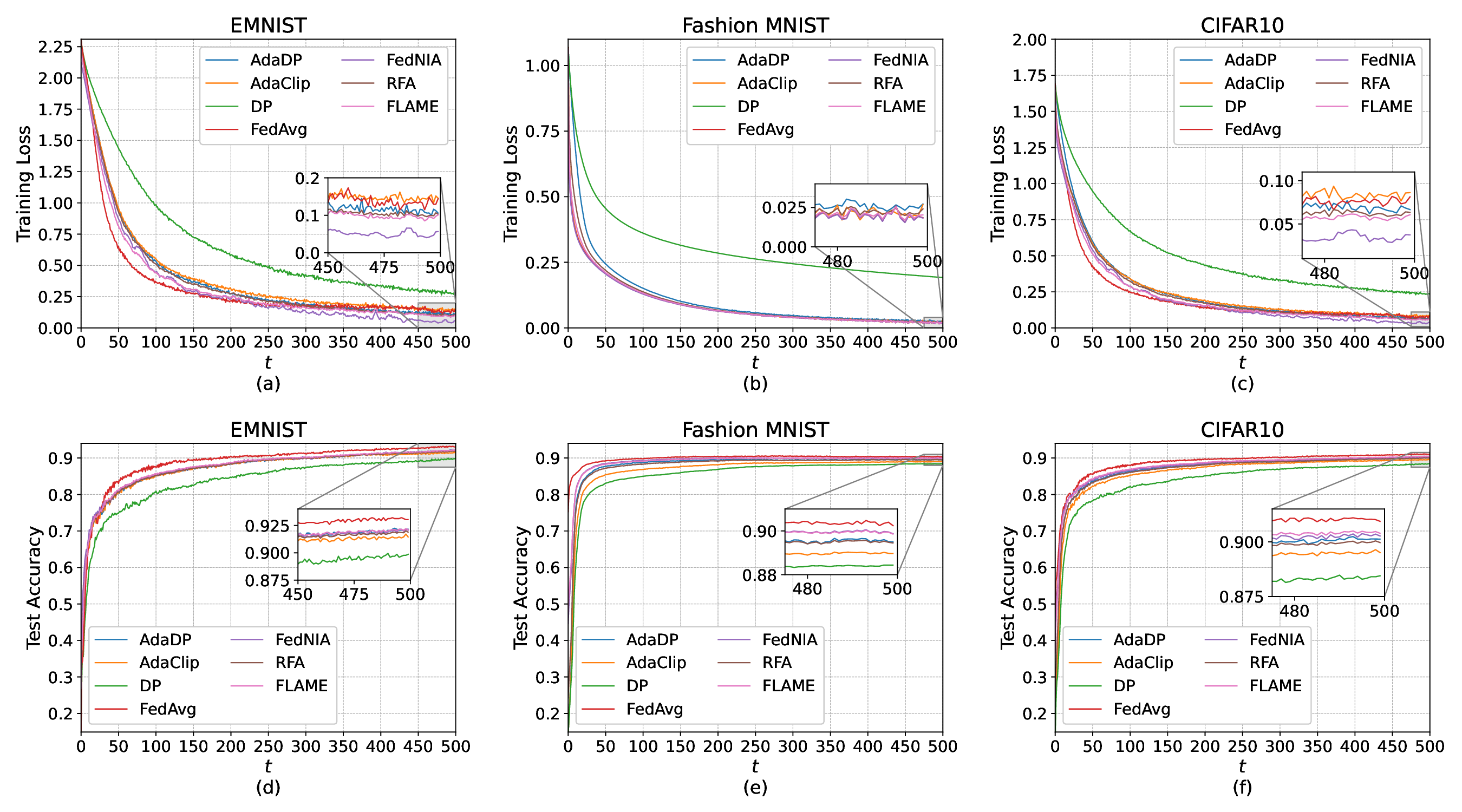}
    \caption{Performance of different aggregation techniques trained on benign data. 100 clients participated in the training process. $t$ denotes the FL iteration number.}
    \label{fig:bengin}
\end{figure*}

\subsection{Experimental Setup}

\subsubsection{Dataset}
The proposed method is evaluated on the Extended MNIST (EMNIST) \cite{7966217}, Fashion-MNIST \cite{xiao2017fashionmnist}, and CIFAR10 \cite{cifar} datasets. To create non-i.i.d. local datasets, a Dirichlet distribution \cite{Minka2000} is used to partition the data among clients with the distribution hyperparameter set to 0.5. The data is batched with 20 samples for EMNIST and Fashion-MNIST, whereas a batch size of 64 is used for CIFAR-10.

\subsubsection{Attacks} 
Attacks are implemented using different ratios of $\delta = \frac{r}{k+r}$, where $r$ and $k$ denote the number of malicious and benign clients, respectively. We consider $\delta$ values in the range $0.02 \leq \delta \leq 0.2$, while keeping the total number of clients fixed at $k + r = 50$. To maximize the effectiveness of attacks, sample poisoning and label flipping are configured with $\gamma = 1$. Untargeted attacks are performed using both sample poisoning and label flipping. For targeted attacks, we simulate targeted label poisoning and backdoor injection. For MNIST, targeted label flipping changes labels 1 and 2 to 7 and 5, respectively. The backdoors are injected into class 1 samples using a fixed pixel pattern trigger $\epsilon$ in the lower right corner. For Fashion-MNIST, label flipping changes class 9 (ankle boot) to 7 (sneaker) and class 0 (T-shirt/top) to 2 (pullover), targeting visually or semantically similar items. Backdoor attacks are applied to samples from class 0 (T-shirt/top) by injecting a visible pixel trigger $\epsilon$ in a fixed location. For CIFAR-10, label flipping is performed on semantically related pairs, such as automobile $\rightarrow$ truck and cat $\rightarrow$ dog. Backdoor attacks are carried out by injecting a small colored square trigger $\epsilon$ into samples from a source class (e.g., airplane) to induce misclassification into a target class (e.g., bird).

\subsubsection{Federated Learning Structure}
For experiments with the EMNIST and Fashion-MNIST datasets, the model structure used for FL is a feedforward neural network with three hidden layers with ReLU activations and sizes 256, 256, and 128. The final layer uses a Softmax activation. Local models undergo five training epochs in each FL round, both clients and the server use Stochastic Gradient Decent (SGD) optimizers. For these experiments, the local learning rate is set to $1$ and $0.02$, respectively. For experiments on CIFAR-10, we use a ResNet-18 \cite{7780459} architecture adapted for $32\times32$ image inputs. The network uses a 3×3 initial convolution (stride 1) without max-pooling, followed by four stages of residual blocks: $[\operatorname{Conv}(64), \operatorname{Conv}(64)] \times 2$, $[\operatorname{Conv}(128), \operatorname{Conv}(128)] \times 2$, $[\operatorname{Conv}(256), \operatorname{Conv}(256)] \times 2$, and $[\operatorname{Conv}(512), \operatorname{Conv}(512)] \times 2$. A global average pooling layer is applied before the final fully connected layer with 10 output neurons. In addition, the local learning rate is set to 0.1 for CIFAR10. In all cases, the number of FL training rounds is set to $T=500$.

\subsubsection{Detector} The detector uses ReLU activations and is optimized using SGD with a learning rate of $0.02$. Encoder layers are set to half of the corresponding layer size of the global model in the first hidden layer, and divided by two for each subsequent layer. The decoder part mirrors this structure with two hidden layers on each side and connects to the encoder via a code layer. The number of generated noise inputs is set to $\nu=100$, and the detector is trained for $\beta = 50$ epochs with a batch size of 10 in each FL round. The parameter $\lambda$ is chosen empirically with the range $[1, 3]$ and set to $\lambda=2$ to provide stable performance across different experiments. Noise inputs are generated from a Gaussian distribution with the same dimensionality as the model input.

\begin{figure*}[t]
    \centering
    
    \includegraphics[trim={0.4cm 0.5cm 0.2cm 0.4cm},clip,width=\textwidth]{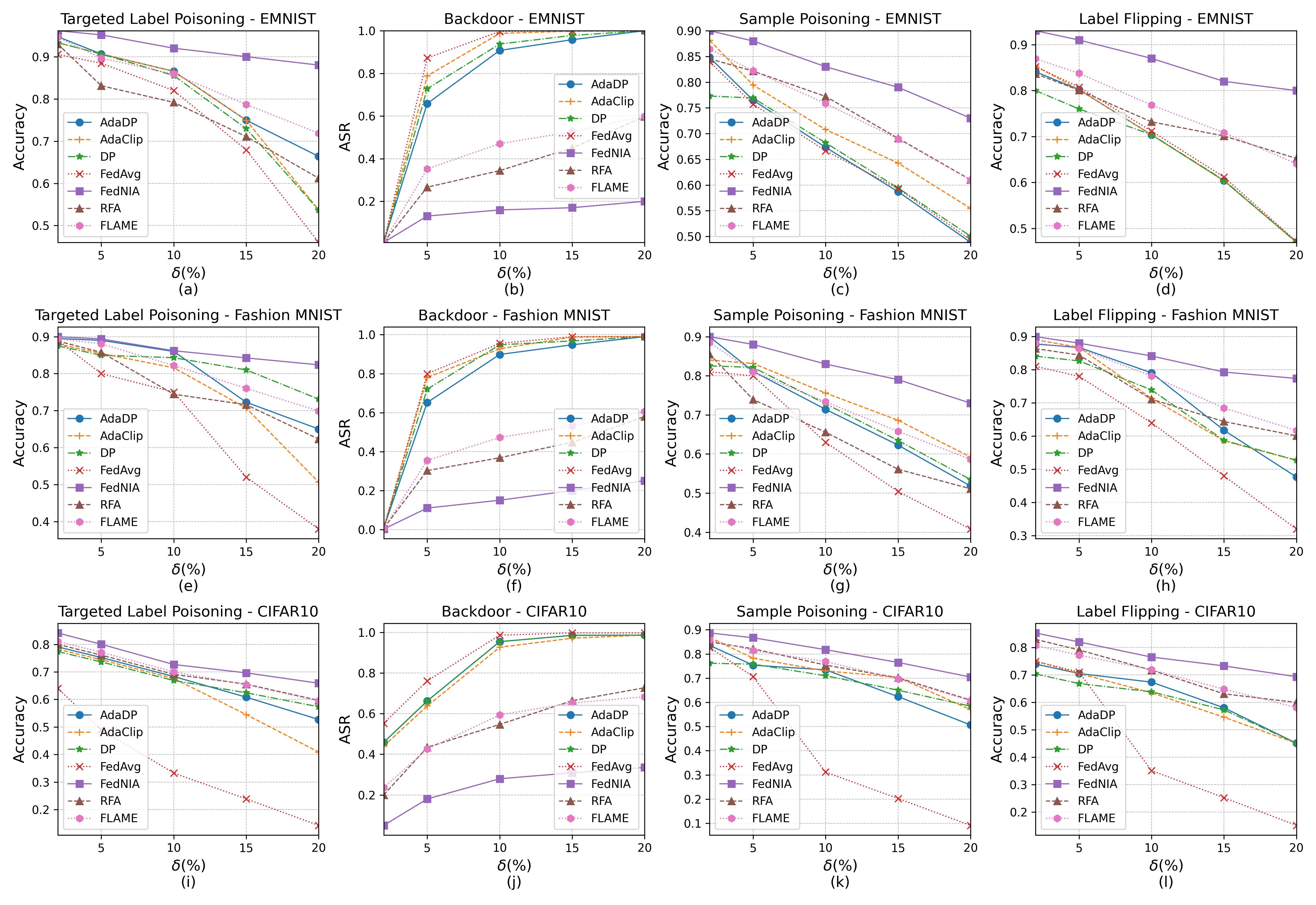}
    \caption{Evaluation of aggregation mechanisms when the FL system is subjected to poisoning attacks. $\delta$ denotes the ratio of attackers to the total number of clients.
    }
    \label{fig:attacks}
\end{figure*}

\subsection{Aggregation Performance}
Before evaluating the selected algorithms under poisoning attacks, we assess their performance when the FL network is not under attack and $\delta=0$. Fig. \ref{fig:bengin} depicts the classification performance using various aggregation techniques when the FL model is trained by 100 benign clients. Comparing the training loss of these algorithms in Fig. \ref{fig:bengin}(a, c), it can be observed that \verb+FedNIA+ initially exhibits a slower convergence speed compared to \verb+FedAvg+; however, given enough time, it outperforms all selected aggregators. This delay in the convergence is due to the fact that the autoencoder network inside \verb+FedNIA+ itself requires convergence and training. Nonetheless, when the autoencoder is finally trained, \verb+FedNIA+ will boost the convergence speed faster than others. Therefore, when the model is not under attack, \verb+FedNIA+ can maintain the model performance and improve the training loss. Similarly, the results in Fig. \ref{fig:bengin}(b) show that \verb+FedNIA+ converges at a speed similar to that of \verb+FedAvg+. However, the convergence speeds in experiments using Fashion MNIST are mostly on par with each other.

The recorded test accuracy during the FL training verifies the previous analysis, as shown in Fig. \ref{fig:bengin}(d). Although the test accuracy of \verb+FedNIA+ initially lags behind \verb+FedAvg+, given enough training time, it reaches an accuracy on par with that of \verb+FedAvg+. Over the full course of training, the proposed method shows consistently better test accuracy than the other aggregators in our experiments. This analysis is also confirmed in Fig. \ref{fig:bengin}(e, f), which depicts the test accuracy on the Fashion MNIST and CIFAR10 datasets.

\subsection{Resilience to Attacks}
The goal of attackers in untargeted attacks is generally to deteriorate the overall accuracy of the FL model. Therefore, sample poisoning and untargeted label flipping are evaluated based on the test accuracy. On the other hand, backdoor and targeted label flipping aim at specific classes. Thus, a test subset is created by drawing the test samples associated with the targeted class and used for calculating the test accuracy. For backdoor attacks, we create a test subset containing samples with triggers and the expected labels that the attacker selects so that the Attack Success Rate (ASR) can be estimated.

\subsubsection{Targeted Data Poisoning} Fig. \ref{fig:attacks} reports the aggregation results for different ratios of $\delta$ and attack mechanisms. As expected, increasing $\delta$ significantly affects the model performance using other aggregators. Nonetheless, \verb+FedNIA+ maintains its robustness to a large extent as $\delta$ varies. The results indicate that \verb+FedNIA+ specifically sets itself apart when it comes to targeted attacks. As shown in Fig. \ref{fig:attacks}(a, b), targeted attacks (i.e., targeted label poisoning and backdoor) are properly mitigated by the proposed aggregation method. In contrast to other methods, this attack almost changes the \verb+FedNIA+ performance with a linear pattern, while slightly changing its accuracy. Moreover, the results of backdoor attacks indicate that even DP-based methods that induce noise in the network weights, still fail to mitigate backdoors when they are initiated by several malicious nodes. Nonetheless, \verb+FedNIA+ sets itself apart in detecting backdoors and significantly outperforms the rest of the methods. It is worth mentioning that the designed backdoors are so subtle that launching them by a single attacker node hardly has any effect on the global state, even on \verb+FedAvg+. Fig. \ref{fig:attacks}(e, f) and Fig. \ref{fig:attacks}(i, j) depict the results of targeted attacks for Fashion MNIST and CIFAR10 datasets, respectively. Similar to the previous analysis, \verb+FedNIA+ outperforms the rest of the methods in terms of accuracy. In addition, this method exhibits more robustness against different ratios of $\delta$. While the comparison between different methods is mostly similar for this case, what stands out is the bigger difference between the baseline performance (\verb+FedAvg+) and the rest of the methods for targeted label poisoning.

\subsubsection{Untargeted Data Poisoning}
The evaluation results for untargeted poisoning attacks are illustrated in Fig. \ref{fig:attacks}(c, d). \verb+FedNIA+ demonstrates high robustness against sample poisoning compared to the other algorithms using EMNIST dataset. However, the difference in the accuracy among the selected aggregators is less pronounced for this attack. As the ratio of malicious clients 
($\delta$) increases, the accuracy of \verb+FedNIA+ remains relatively stable, showcasing its effectiveness in mitigating the adverse effects of data poisoning. Furthermore, \verb+FedNIA+ effectively eliminates various ratios of the label flipping attack, maintaining relatively high performance. In contrast, the accuracy of the other methods drastically deteriorates as $\delta$ increases. Looking at Fig. \ref{fig:attacks}(g, h) and Fig. \ref{fig:attacks}(k, l), it can be observed that \verb+FedNIA+ also outperforms the rest of the methods when tested on Fashion MNIST and CIFAR10 datasets. These results also confirm the findings from EMNIST experiments. The most noticeable difference in this set of results is the smaller difference between the performance of \verb+FedNIA+ and the rest, albeit still significant.

\subsection{Adaptive Adversaries}
\begin{figure}[t]
\centering

\includegraphics[trim={0.2cm 0.2cm 0.2cm 0.2cm},clip,width=0.8\columnwidth]{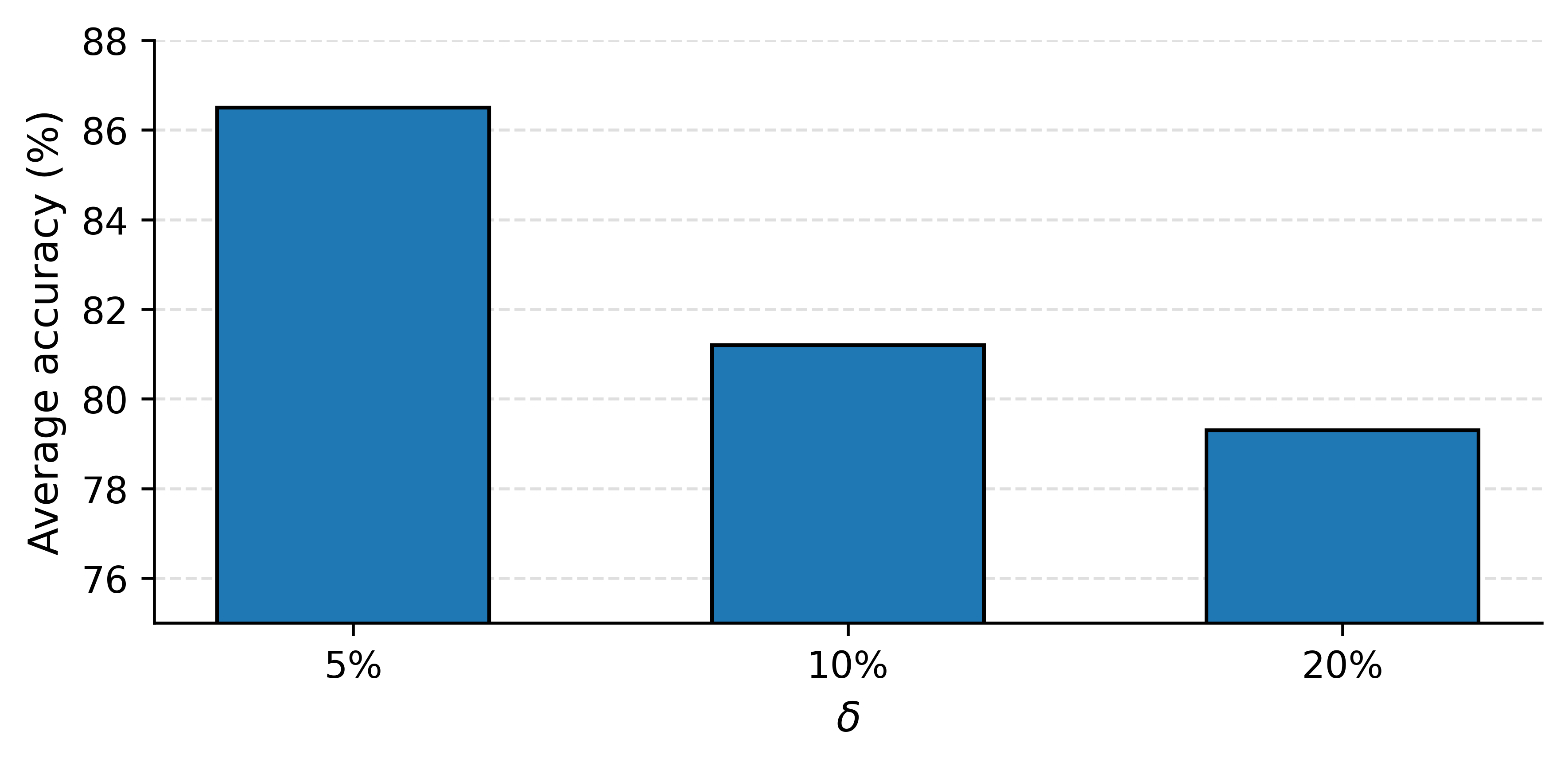}
\caption{Average accuracy of \texttt{FedNIA} under black-box adaptive adversaries across all utilized datasets.}
\label{fig:adaptive}
\end{figure}

To further evaluate \verb+FedNIA+ against adaptive adversaries, we consider a
black-box adaptive attack under the same threat model \cite{11463993}, where malicious
clients have access to their local models but no knowledge of the
server-side defense mechanism. As shown in Fig.~\ref{fig:adaptive},
\verb+FedNIA+ maintains relatively stable classification performance as the
ratio of adaptive adversaries increases, demonstrating its resilience
under this tougher black-box attack setting.

\subsection{Runtime Analysis}
\label{sec:runtime}

\begin{figure}[t]
    \centering
    
    \includegraphics[trim={0.3cm 0.5cm 0.3cm 0.3cm},clip,width=0.85\columnwidth]{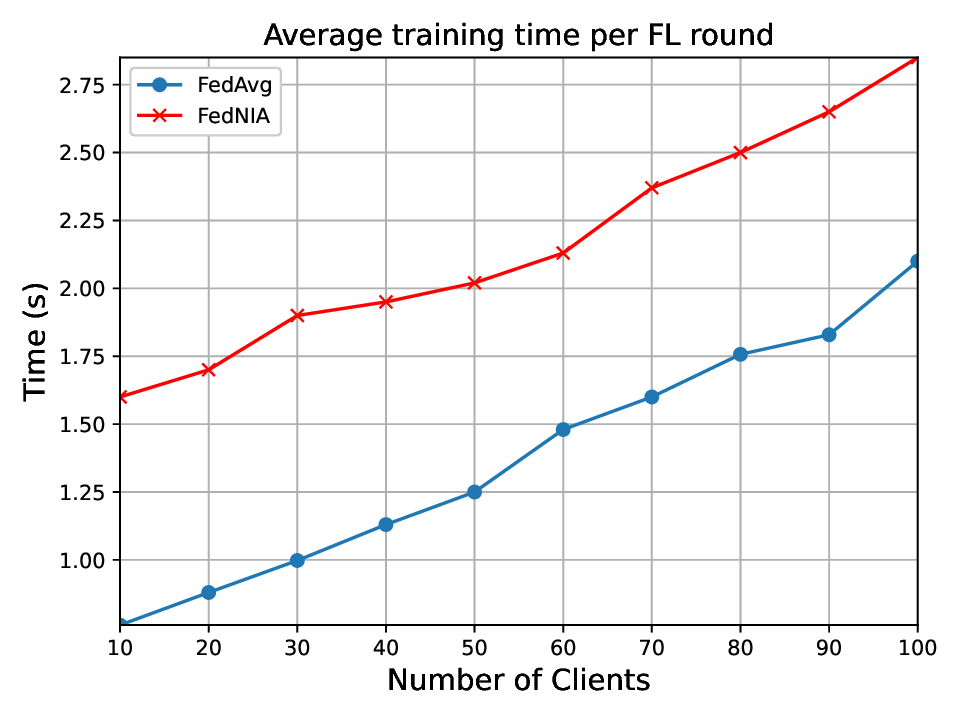}
    \caption{Averaged runtime comparison between the proposed method and the baseline, FedAvg. The time is recorded for one FL training round and averaged for EMNIST, Fashion MNIST, and CIFAR10 experiments.}
    \label{fig:time}
\end{figure}

Fig. \ref{fig:time} presents a comparison of the average runtime for a single FL training iteration using each of the selected aggregators. The recorded run time is averaged for both datasets and all attacks and ratios. These measurements were obtained on a computer equipped with an NVIDIA RTX 3080 GPU, an Intel Core i7-12700 CPU, and 32 GB of RAM. The experiments were conducted using TensorFlow Federated on an Ubuntu kernel, accessed via the Windows Subsystem for Linux. The results suggest that the superior balance between accuracy and security offered by \verb+FedNIA+ comes with a higher computational cost, a common trait among robust aggregators. However, similar to \verb+FedAvg+, the runtime of \verb+FedNIA+ is linearly correlated with the total number of clients. This finding aligns with the previous complexity analysis of the proposed algorithm. While we consider the enhanced robustness worth the increased computational cost, future research could explore ways to improve the algorithm's efficiency.

\begin{table*}[t]

\centering
\caption{Friedman test p-values, effect sizes, and rank sums per dataset (Lower rank sum is better). }
\begin{tabular}{llccccccccc}
\toprule
Attack & Dataset & \verb+FedNIA+ & \verb+FLAME+ & \verb+RFA+ & \verb+AdaDP+ & \verb+AdaClip+ & \verb+DP+ & \verb+FedAvg+ & p-value & Effect Size\\
\midrule

\multirow{3}{*}{Targeted Label Poisoning}
& EMNIST  & 5 & 16 & 30 & 12 & 20 & 24 & 33 & $3.02\times10^{-4}$ & 0.84 \\

& Fashion-MNIST & 5 & 16 & 26 & 14 & 28 & 20 & 31 & $1.59\times10^{-3}$ & 0.71\\

& CIFAR10  & 5 & 12 & 13 & 22 & 27 & 26 & 35 & $9.63\times10^{-5}$ & 0.93 \\

\midrule

\multirow{3}{*}{Backdoor}
& EMNIST  & 5 & 15 & 13 & 21.5 & 30 & 27 & 28.5 & $6.99\times10^{-4}$ & 0.78\\

& Fashion-MNIST  & 5 & 15 & 12 & 22.5 & 27.5 & 28 & 30 & $6.93\times10^{-4}$ & 0.78\\

& CIFAR10  & 5 & 12 & 13 & 27 & 21 & 27 & 35 & $5.56\times10^{-5}$ & 0.97\\

\midrule

\multirow{3}{*}{Sample Poisoning}

& EMNIST& 5 & 14 & 14 & 30 & 18 & 27 & 32 & $2.81\times10^{-4}$ & 0.85\\

& Fashion-MNIST  & 5 & 17 & 29 & 21 & 13 & 21 & 34 & $5.04\times10^{-4}$ & 0.80\\

& CIFAR10 & 5 & 15 & 14 & 27 & 18 & 27 & 34  & $3.37\times10^{-4}$ & 0.83 \\

\midrule

\multirow{3}{*}{Label Flipping}

& EMNIST  & 5 & 11 & 20 & 27 & 28 & 30 & 19  & $1.07\times10^{-3}$ & 0.74\\

& Fashion-MNIST  & 5 & 14 & 22 & 19 & 20 & 25 & 35 & $1.15\times10^{-3}$ & 0.74\\

& CIFAR10  & 5 & 13 & 12 & 25 & 26 & 29 & 30  & $3.11\times10^{-4}$ & 0.84\\

\bottomrule
\end{tabular}
\label{tab:stats}
\end{table*}

\subsection{Significance Test}

\begin{figure}[t]
        \centering
        \includegraphics[trim={0.3cm 0.6cm 0.3cm 0.3cm},clip,width=0.85\columnwidth]{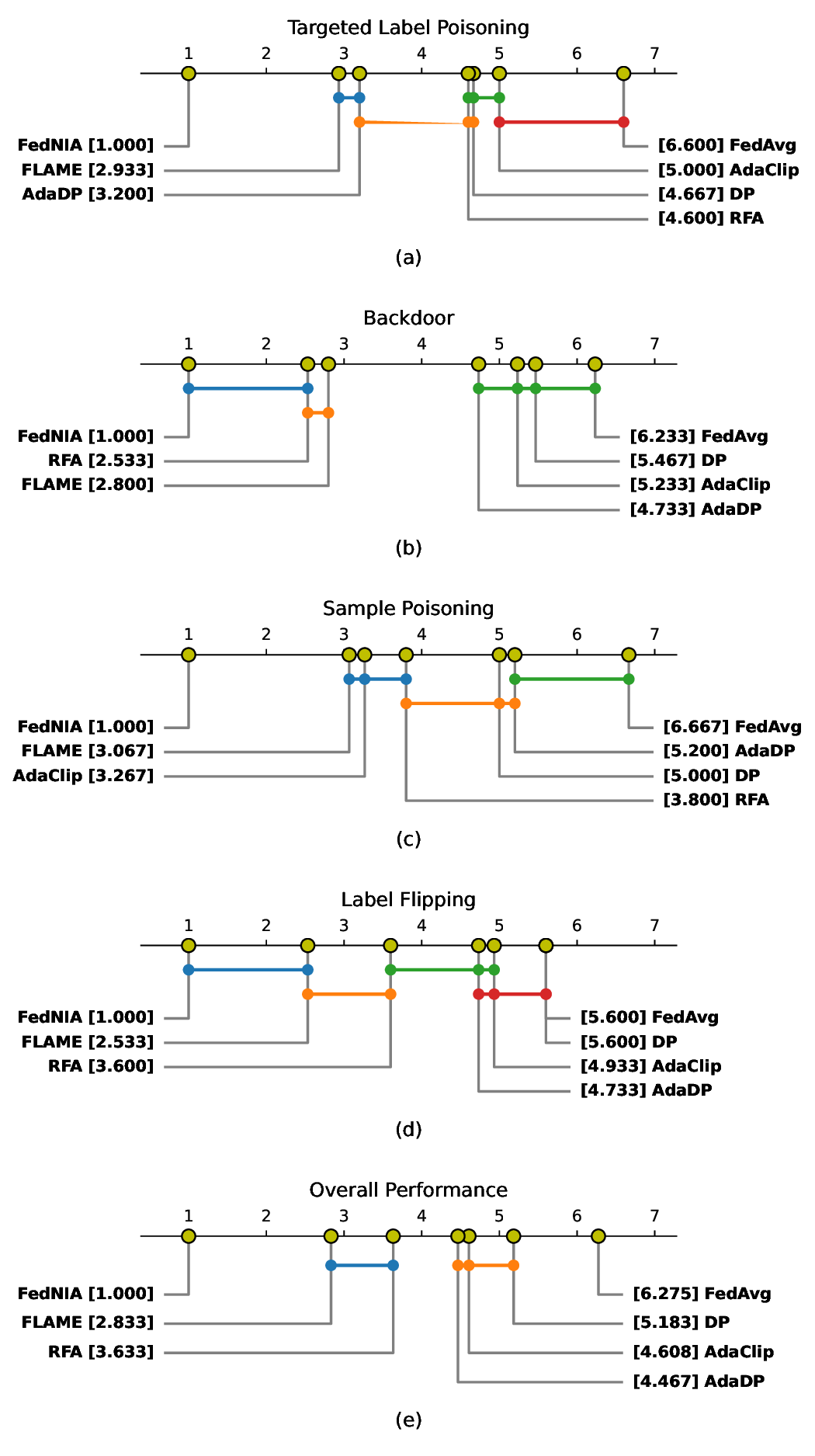}
    \caption{Critical difference diagram obtained from the post-hoc Friedman test.} 
    \label{fig:ttest}
\end{figure}

Fig.~\ref{fig:ttest} presents the Critical Difference (CD) diagrams derived from the post-hoc Friedman test, while the corresponding p-values, effect sizes, and rank sums across datasets are reported in Table~\ref{tab:stats}. The Friedman test evaluates whether the observed performance differences among methods are statistically significant across multiple experiments. Methods that are not significantly different are connected in the CD diagrams, indicating statistical similarity. Across all attack types and datasets, the consistently low p-values (ranging from $10^{-3}$ to $10^{-5}$) confirm statistically significant differences among the compared methods, while the high effect sizes indicate strong agreement in their relative rankings. In terms of performance, \verb+FedNIA+ consistently achieves the lowest rank sum across all settings, demonstrating stable and superior performance across datasets. From the CD diagrams, several methods exhibit statistical similarity in specific scenarios. For example, under backdoor and label flipping attacks (Fig.~\ref{fig:ttest}(b, d)), \verb+FedAvg+, \verb+AdaClip+, and \verb+DP+ are often grouped, indicating comparable performance. Similarly, in sample poisoning and overall performance scenarios (Fig.~\ref{fig:ttest}(c, e)), \verb+AdaDP+, \verb+AdaClip+, and \verb+DP+ show overlapping groups, suggesting similar behavior. The only cases where competing methods approach \verb+FedNIA+ occur under backdoor and label flipping attacks, where \verb+FedNIA+ remains top-ranked but is grouped with \verb+RFA+ and \verb+FLAME+, respectively. However, in the majority of settings, \verb+FedNIA+ is statistically separated from other methods, indicating a consistent and significant performance advantage. These results collectively demonstrate the robustness and generalization capability of the proposed method against diverse data poisoning attacks in FL.

\subsection{Overall Performance}

\begin{figure}[t]
\centering

\includegraphics[width=0.8\columnwidth, trim={0.6cm 0.1cm 1.2cm 1cm},clip]{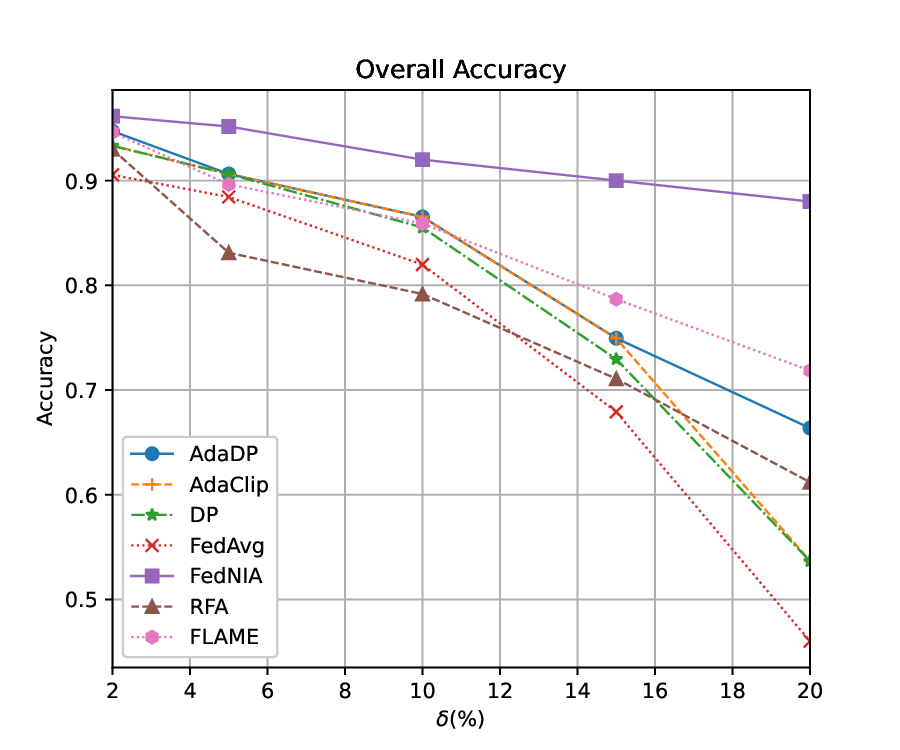}
\caption{Overall average accuracy of aggregation methods computed across all evaluated datasets.}
\label{fig:overall}
\end{figure}

Fig. \ref{fig:overall} shows the overall accuracy of robust aggregators over all experiments. The results are averaged for all attacks and datasets. For Backdoor attacks, we converted ASR to accuracy by calculating $1-\operatorname{ASR}$. As indicated in this figure, \verb+FedNIA+ outperforms all when considering the overall performance. Moreover, \verb+FedAvg+ results in the lowest performance when under attack, as expected. The rest of the methods exhibit a somewhat similar performance, especially in higher $\delta$ ratios. Nonetheless, as explained before, this behavior may not be always the case when studying certain attacks and datasets, as these results only reflect the overall performance of each method throughout the experiments.

Table~\ref{tab:ae} reports the overall generalization and stability of the autoencoder across FL rounds. The consistently small reconstruction error gap, computed using disjoint noise inputs, indicates that the autoencoder generalizes to unseen activation patterns and does not overfit to transient global states. Moreover, the low standard deviation of the global loss across runs confirms that filtering suspicious updates does not destabilize convergence and results in a stable training process.

\begin{table}[t]

\centering
\caption{$AE$ generalization and stability across FL rounds. The error gap measures the relative difference between train and test reconstruction errors on noise-induced activations. Average standard deviation (Avg. std.) denotes the standard deviation of global loss across multiple runs, averaged over rounds.}
\begin{tabular}{lcc}
\toprule
Dataset	& Error Gap (\%) & Avg. Std. (\%) \\
\midrule
EMNIST & 3.25 & 0.15 \\
Fashion-MNIST & 3.81 & 0.24  \\
CIFAR10 & 5.62 & 0.37 \\
\bottomrule
\end{tabular}
\label{tab:ae}

\end{table}

\subsection{Threshold Sensitivity}

To further evaluate the thresholding strategy in \verb+FedNIA+, we analyze the
effect of the sensitivity parameter $\lambda$ in
$\tau=\mu+\lambda\sigma$. We vary $\lambda\in\{1,2,3\}$ and evaluate
the resulting false alarm rate, corresponding to benign client
models incorrectly filtered, and missed detection rate,
corresponding to malicious client models that remain undetected.

\begin{figure}
\centering

\includegraphics[trim={0.2cm 0.3cm 0.2cm 0.3cm},clip,width=0.8\columnwidth]{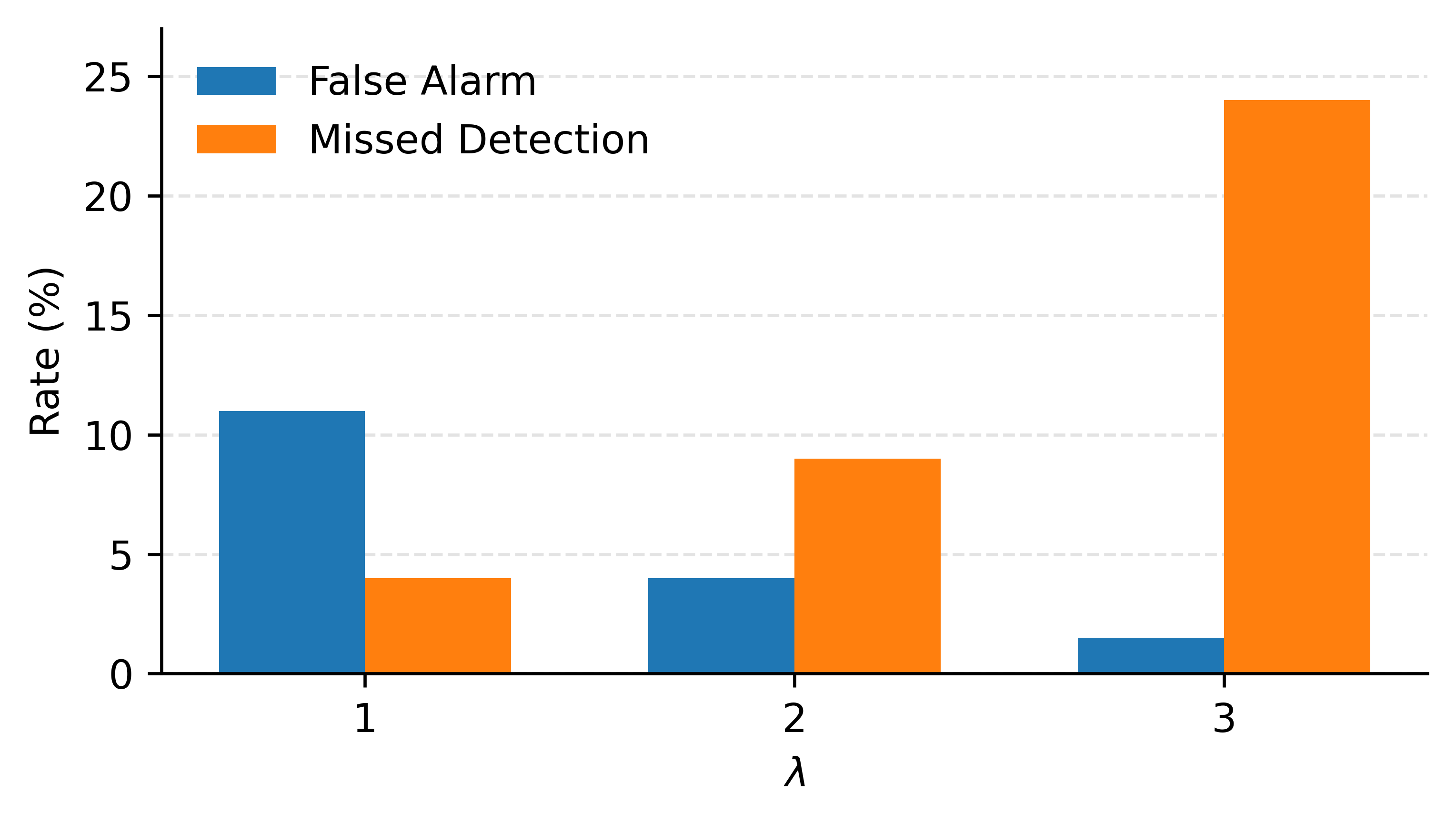}
\caption{Effect of the threshold parameter $\lambda$ on false alarm and missed detection rates.}
\label{fig:lambda}

\end{figure}

As shown in Fig.~\ref{fig:lambda}, $\lambda$ controls the trade-off
between the two types of detection error. A smaller value of $\lambda$
results in a lower threshold and more aggressive filtering, reducing
missed detections at the cost of an increased false alarm rate. In
contrast, increasing $\lambda$ makes the filtering criterion more
conservative, substantially reducing false alarms while increasing
missed detections. The intermediate setting $\lambda=2$ provides a
favorable balance between these competing effects, considerably
reducing false alarms relative to $\lambda=1$ without the substantial
increase in missed detections observed at $\lambda=3$. Therefore,
$\lambda=2$ is adopted as the default value throughout the experiments.
These results provide empirical support for the selected threshold
setting and demonstrate its expected sensitivity behavior.

\section{Conclusion}
\label{sec:conclusion} 
This paper contributes to the ongoing efforts to secure FL systems against adversaries by presenting \verb+FedNIA+, a novel mechanism to fortify FL systems against a wide array of data poisoning attacks. This method is unique in its ability to address both targeted and untargeted attacks, including sample poisoning, label flipping, and backdoor attacks that are launched by several malicious nodes. By injecting random noise inputs into the reconstruction of client models and analyzing the layer activations, \verb+FedNIA+ was able to identify and exclude malicious clients before the aggregation process. In addition, \verb+FedNIA+ eliminates the need for a central test dataset for client evaluation, a common yet often impractical assumption in existing defense mechanisms. Experimental results demonstrate the effectiveness of our approach. The proposed method not only successfully identifies and excludes malicious updates, but also maintains a performance level similar to the baseline when no attack is present. This balance between resilience and performance in the absence of threats is a significant advantage of the proposed approach. Moreover, \verb+FedNIA+ has demonstrated strong empirical performance across sample poisoning, label flipping, and backdoor attacks, covering the primary categories of data poisoning attacks when federated data is non-i.i.d.

In addition to computational overhead, which may limit applicability on resource-constrained devices, \verb+FedNIA+ may be challenged by more advanced adaptive adversaries and extreme data imbalance conditions. Examining these scenarios in depth requires dedicated investigation and is beyond the scope of the current study, and is therefore identified as an important direction for future research. Thus, future work will focus on improving efficiency and investigating robustness under these conditions as well as extending the approach to support other FL paradigms beyond model weight sharing.

\bibliography{main.bib}
\bibliographystyle{IEEEtran}

\end{document}